\documentclass[sigconf,natbib=false]{acmart}
\AtBeginDocument{%
  }

\usepackage{booktabs}
\usepackage{multirow}
\usepackage{adjustbox}
\usepackage{graphicx}
\usepackage{makecell}
\usepackage[normalem]{ulem}
\setcopyright{acmlicensed}
\copyrightyear{2026}
\acmYear{2026}
\setcopyright{cc}
\setcctype{by}
\acmConference[MM '26]{Proceedings of the 34th ACM International Conference on Multimedia}{November 10--14, 2026}{Rio de Janeiro, Brazil}
\acmBooktitle{Proceedings of the 34th ACM International Conference on Multimedia (MM '26), November 10--14, 2026, Rio de Janeiro, Brazil}
\acmDOI{10.1145/3767308.3838661}
\acmISBN{979-8-4007-2213-4/2026/11}

\RequirePackage[
  datamodel=acmdatamodel,
  style=acmnumeric,
  ]{biblatex}

\begin{document}

\title{Neural video codecs quality assessment dataset and benchmark}


\author{Nickolay Safonov}
\affiliation{%
  \institution{AI Center, Lomonosov Moscow State University}
  \city{Moscow}
  \country{Russia}}
\affiliation{%
  \institution{MSU Institute for Artificial Intelligence}
  \city{Moscow}
  \country{Russia}}
\email{nikolay.safonov@graphics.cs.msu.ru}

\author{Nikita Gornostaev}
\affiliation{%
  \institution{Lomonosov Moscow State University}
  \city{Moscow}
  \country{Russia}}
\email{nikita.gornostaev@graphics.cs.msu.ru}

\author{Alexandra Dubonos}
\affiliation{%
  \institution{Lomonosov Moscow State University}
  \city{Moscow}
  \country{Russia}}
\email{alexandra.dubonos@graphics.cs.msu.ru}

\author{Dmitriy S. Vatolin}
\affiliation{%
  \institution{MSU Institute for Artificial Intelligence}
  \city{Moscow}
  \country{Russia}}
\email{dmitriy@graphics.cs.msu.ru}


\renewcommand{\shortauthors}{N. Safonov, N. Gornostaev, A. Dubonos, \& D. Vatolin}

\begin{abstract}
 Video traffic constitutes a significant share of global web traffic. To reduce its volume, video codecs have been developed and continuously improved. While the industry has achieved substantial progress in traditional video coding, neural video codecs (NVCs) have recently emerged as a new approach that applies deep learning to video compression. This creates new challenges for compression quality assessment, which is essential for the further development and improvement of such codecs. In particular, it is important to evaluate the novel temporal compression paradigms introduced by NVCs. In this work, we present a large-scale subjective dataset of videos compressed with both neural and traditional video codecs. The subjective scores were collected through crowd-sourced pairwise comparisons. The proposed dataset provides a valuable resource for the development and benchmarking of video quality metrics tailored to neural video codecs. The dataset is available at the following link: \url{https://videoprocessing.github.io/nvc-dataset-benchmark}.
\end{abstract}




\maketitle
\section{Introduction}
\begin{table*}[t]
  \caption{Summary of subjective compressed video quality datasets including neurally compressed and the proposed dataset.}
  \label{tab:freq}
  \centering
  \resizebox{\textwidth}{!}{%
  \begin{tabular}{llcccccc}
    \\[0.5em]
    \toprule
    &Dataset &Source &Orig. &Dist. &Subjective Framework &Subj. &Ans. \\
    \midrule
    \multirow{17}{*}{\textbf{General}} 

    &MCL-JCV (2016)~\cite{wang2016mcl}  &Raw &30  &1,560  &In-lab &150 &78K \\
    &VideoSet (2017)~\cite{wang2017videoset} &Raw &220 &45,760 &In-lab &800 &- \\
    &SJTU-4K (2017)~\cite{zhu2016sjtu}   &Raw &20  &200   &In-lab &30  &6K \\
    &GamingVSET (2018)~\cite{barman2018gamingvideoset} &Raw &24  &576   &In-lab &25  &- \\
    &NFLX (2016)~\cite{li2016toward}   &Raw &12  &300   &In-lab &54  &9K \\
    &KUGVD (2019)~\cite{barman2019no} &Raw &6   &144   &In-lab &17  &- \\

    &UGC-VIDEO (2020)~\cite{li2020ugc} &UGC &50  &550   &In-lab &30  &16.5K \\
    &AVT-VQDB (2019)~\cite{rao2019avt} &UGC &15  &300   &In-lab &50  &15K \\
    &TGV (2022)~\cite{wen2022subjective} &UGC &150 &1,143 &In-lab &19  &- \\
    &TaoLive (2023) ~\cite{zhang2023md} &UGC &418  &3,762 &In-lab &44 &165.5K \\
    &KVQ (2024) ~\cite{lu2024kvq} &UGC &600 &4,200 &In-lab &15 &63K \\

    &CVQAD (2022)~\cite{antsiferova2022video} &Raw+UGC &36  &1,022 &Crowd. &10,800 &320K \\
    &LEHA-CVQAD (2025)~\cite{gushchin2025leha} &Raw+UGC &59 &6,240 &Crowd. &11,000 &400K \\

    &YT-UGC+ (2021) ~\cite{wang2021rich} &UGC &189 &567  &In-lab &30 &17K \\
    &HDR-sport (2023) ~\cite{shang2023subjective} &Raw &12 &42 &In-lab &140 &32K \\
    &BrightVQA (2025)~\cite{brightvq2025} &UGC &300 &2100 &Crowd. &200 &74K \\
    &AVT-VQDB-UHD-1-HDR(2024)~\cite{rao2024avt} &Raw &5 &195 &In-lab &24 &4.7K\\
    &Shang2022 (2022)~\cite{shang2022subjective} &Raw &31 &310 &In-lab &66 &22K \\ 
    &SCDB (2025)~\cite{safonov2025screen} &Raw &100 &1,600 &Crowd. &8,000 &120K \\ 

    \midrule
    \multirow{3}{*}{\textbf{Neural Compression}} 

    &AVT-VQDB-UHD-1-NVC (2025)~\cite{herb2025evaluating} &RAW &6  &216   &In-lab &30  &6.5K \\
    &CLIC (2024)~\cite{clic2024_tasks} &Raw+UGC &62  &1,260 &In-lab &- &- \\

    &\textbf{Proposed} &Raw+UGC &80 &2,880 &Crowd. &12,000 &504K \\

    \bottomrule
  \end{tabular}}
\end{table*}

Video streaming represents the dominant portion of global internet traffic. As data delivery costs continue to rise, providers are required to improve compression efficiency while preserving perceptual quality, motivating ongoing advancements in video codec design.
Neural video codecs (NVCs), which apply deep learning techniques to video compression, have recently emerged as a promising alternative to traditional codecs. This trend is driven by the increasing demand for efficient compression of complex visual content in applications such as streaming, videoconferencing, and user-generated media. Unlike conventional video coding, NVCs rely on learned representations and often introduce new temporal compression strategies. As a result, the characteristics of compression artifacts and distortions differ significantly from those produced by traditional codecs. This creates new challenges for maintaining and assessing visual quality, as existing evaluation methods may not adequately capture the perceptual effects introduced by neural compression.
The optimization of video codecs fundamentally depends on video quality assessment (VQA) metrics, which guide parameter selection by quantifying the trade-off between bitrate and perceived visual quality. VQA metrics are generally categorized into full-reference (FR) methods, which require access to the original undistorted video, and no-reference (NR) methods, which operate without such information. Alongside traditional metrics such as PSNR, SSIM, and VMAF, several modern learning-based metrics have been introduced in recent years. While these metrics often perform well for traditional codecs, they have not been systematically evaluated for neural video codecs (NVCs), which introduce distortions that differ from those of conventional encoders, including novel temporal artifacts. VQA metrics trained on natural content may therefore fail to generalize to NVC-compressed videos and can produce inaccurate quality predictions. Additionally, it should be noted that some codecs adopt generative approaches, where part of the visual information is synthesized at the decoder. This introduces new challenges for quality assessment, as the reconstructed content may deviate from the original while remaining perceptually plausible, making accurate quality estimation more difficult. In this work, we evaluate the performance of recent VQA metrics on videos compressed with NVCs and provide a benchmark of current progress in this domain.

We introduce a large-scale subjective dataset for evaluating the perceptual quality of neural video codecs (NVCs), addressing the limited availability of diverse benchmarks in this area. The dataset comprises 2,880 distorted video sequences generated with both neural and conventional codecs. Subjective quality is annotated via crowdsourced pairwise comparisons and aggregated using the Bradley–Terry model~\cite{bradley1952rank}. We further benchmark a diverse set of state-of-the-art VQA metrics, including full-reference baselines and recent no-reference deep models.

Our contributions are as follows:

\begin{itemize}
    \item \textbf{We provide a new large-scale subjective NVC dataset}, containing 2,880 distorted video sequences with 6 neural and 4 classical codecs.
    \item \textbf{We benchmark a diverse set of VQA metrics}, spanning full-reference and no-reference approaches, to evaluate their effectiveness on neural and conventional codecs and to highlight systematic differences in their behavior.
\end{itemize}

By combining diverse compression artifacts, large scale, and comprehensive annotations, our dataset provides a robust benchmark for developing, evaluating, and improving VQA metrics for neural video codecs. It addresses an important gap in the field and supports future research on generalizable and perceptually aligned video quality models.

\section{Related works}

The training and evaluation of video quality assessment models require datasets annotated with human subjective judgments. A number of benchmarks have been proposed that provide such annotations for distorted video content, with an overview of recent datasets summarized in Table~\ref{tab:freq}. While recent work has focused on constructing large-scale datasets capturing compression artifacts from traditional codecs~\cite{wang2016mcl, wang2017videoset, li2020ugc, barman2018gamingvideoset, barman2019no, wen2022subjective, antsiferova2022video}, datasets tailored to screen content remain comparatively underexplored. Due to the fundamental differences between classical and neural compressed videos in terms of visual structure and artifact characteristics, we specifically focus our review on datasets designed for NVC.

Existing subjective video quality datasets can be broadly categorized into three groups: legacy datasets with raw source content, UGC datasets, and hybrid collections that combine both paradigms.

Legacy datasets~\cite{wang2016mcl, wang2017videoset, zhu2016sjtu, barman2018gamingvideoset, li2016toward, barman2019no, shang2023subjective, rao2024avt, shang2023subjective, safonov2025screen} are constructed from pristine reference videos and employ controlled compression pipelines. While they provide high-quality ground truth and well-defined distortions, they exhibit limited content diversity and fail to reflect the complex artifacts encountered in real-world scenarios. In particular, the exclusive use of raw (uncompressed) sources omits distortions introduced by consumer devices, editing pipelines, and platform-specific processing.

In contrast, UGC datasets~\cite{li2020ugc, rao2019avt, wen2022subjective, zhang2023md, lu2024kvq, wang2021rich, brightvq2025} naturally encompass a wide spectrum of authentic distortions arising from diverse capture conditions, codecs, and user behaviors. This makes them highly representative of in-the-wild content. However, the absence of corresponding pristine references limits their applicability for full-reference (FR) VQA methods. Furthermore, UGC videos are typically subject to additional, often unknown, recompression during platform delivery, which complicates both analysis and reproducibility. 

Hybrid datasets~\cite{antsiferova2022video, gushchin2025leha, herb2025evaluating, clic2024_tasks} aim to bridge this gap by combining user-generated content with controlled synthetic distortions. In these datasets, authentic videos are augmented with additional degradations, enabling joint modeling of natural and systematically induced artifacts. While this design improves coverage of the distortion space, some hybrid datasets remain partially or fully inaccessible, limiting their usability for benchmarking and reproducible research.

The Challenge on Learned Image Compression (CLIC) 2024 provides a large-scale benchmark for evaluating neural compression methods under perceptual criteria. Unlike traditional datasets with fixed distortions, the CLIC dataset is dynamically constructed from submissions of participating methods, resulting in a diverse collection of compressed images and videos with realistic artifacts produced by state-of-the-art neural codecs~\cite{clic2024_tasks}. Recent work~\cite{herb2025evaluating} investigates the performance of video quality assessment (VQA) metrics on both neural and traditional codecs using high-resolution 4K/UHD-1 content. The dataset introduced in this study focuses on compression artifacts produced by modern learned video compression methods alongside conventional codecs, enabling a direct comparison of their perceptual characteristics. However, the dataset is rather small.

These limitations highlight the need for large-scale, realistic datasets tailored to neural video compression (NVC). To address this gap, we introduce a dataset specifically designed for evaluating neural compression methods. Table~\ref{tab:freq} shows comparison of proposed dataset features with the existing ones. The dataset enables reliable benchmarking of objective quality metrics under realistic conditions and provides a foundation for developing perceptually aligned evaluation methods for NVC.

\section{Dataset}

In this section, we describe the dataset construction process, including source video selection, generation of compressed sequences, and subjective data collection. The dataset consists of 80 original screen content videos, each 10 seconds long and in 1080p resolution. Each source video is compressed using 10 different codecs, including 6 NVC and 4 traditional, with various rate–distortion settings, resulting in a total of 2,880 distorted sequences. To obtain reliable ground-truth annotations, we collected over 504,000 pairwise subjective comparisons from more than 12,000 unique assessors. The collected votes are aggregated using the Bradley–Terry model~\cite{bradley1952rank} to derive consistent quality scores. The primary objective of this dataset is to provide a robust benchmark for evaluating both full-reference and no-reference VQA models under realistic conditions representative of neural video compression.
\begin{figure*}[h]
  \centering
  \includegraphics[width=\textwidth]{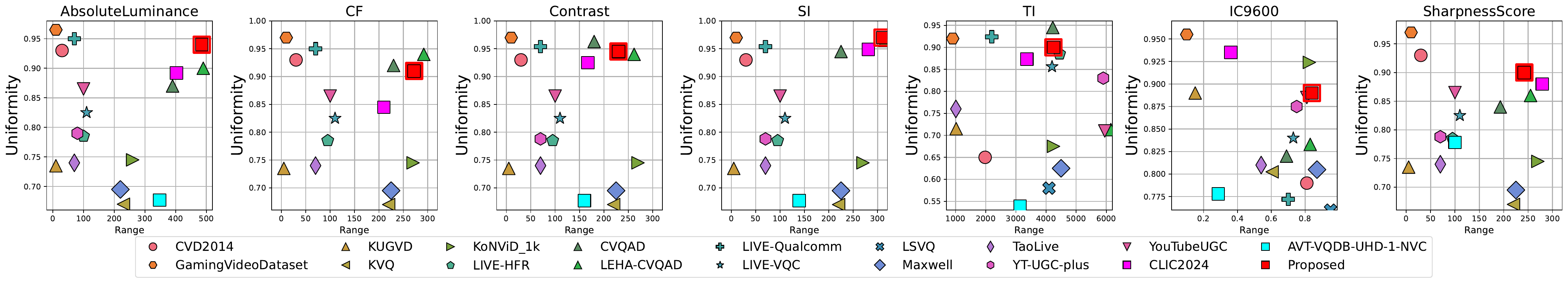}
  \caption{Statistics (Range, Uniformity) of various complexity metrics for proposed and other datasets.}
  \label{fig:guh}
\end{figure*}
\subsection{Video Preparation}
To assemble a representative set of high-quality source videos, we first defined requirements on visual fidelity, diversity of content, and licensing constraints. Candidate material was gathered from publicly available platforms, including Vimeo, Xiph.Org and YouTube UGC, resulting in a large initial pool. To ensure that the source content was not affected by strong pre-existing compression artifacts, only videos with bitrates above 20 Mbps were retained. All selected videos were converted to a unified YUV 4:2:0 format to standardize further processing. Only videos available under permissive licenses (CC BY or CC0) were considered.

We computed spatial and temporal complexity characteristics for all videos in the dataset. Spatial complexity was estimated as the average size of x264-encoded I-frames normalized by the uncompressed frame size, while temporal complexity was defined as the ratio between the average sizes of P-frames and I-frames.

Rather than directly selecting videos from this pool, we aimed to ensure broad and balanced coverage of content characteristics. For this purpose, each candidate video was described using spatial and temporal complexity indicators, capturing variations in motion and structural detail. These descriptors were then used to organize the dataset via K-means clustering into 80 groups, each representing a distinct region of the content space. From every group, a small subset of videos was sampled and manually inspected, after which a single representative example was chosen. This selection process was guided by the goal of achieving both visual quality and diversity across semantic categories.

As a result, the final reference set covers a wide range of scenarios, including dynamic scenes such as sports and gaming, natural environments, talking-head and interview footage, broadcast and animation content, as well as various forms of user-generated material (e.g., vlogs and advertisements). Particular attention was also given to include visually specific patterns such as water surfaces and close-up facial content, which are known to be challenging for compression and quality assessment. Also we have included several screen content videos from the ~\cite{safonov2025screen} collection.

For the distorted video generation, we selected six modern neural video codecs: OpenDVC~\cite{lu2019dvc}, DCVC-RT~\cite{jia2025towards}, NEVC~\cite{liao2025ehvc}, GLC~\cite{qi2025generative}, BRHVC~\cite{liu2025neural}, and DHVC~\cite{lu2024deep}. These codecs represent different approaches to neural compression and produce a variety of artifacts. For each codec, we generated compressed sequences at four rate–distortion settings to cover a wide range of quality levels. This setup allows us to capture diverse compression artifacts, including those specific to neural methods, and enables comparison with traditional codecs under similar conditions. We included four standard codecs corresponding to widely used compression standards H.264/AVC, H.265/HEVC, H.266/VVC, and AV1 to provide a consistent baseline for comparison. The details on codecs presets and settings may be found in the supplementary.

\subsection{Subjective Assessment}

We employed a subjective pairwise comparison protocol to rank the distorted video sequences. Annotations were collected via crowdsourcing using a sequential preference interface. In each trial, participants were presented with two videos one-by-one and asked to indicate which one had higher visual quality. Three response options were provided: \textit{left}, \textit{right}, or \textit{cannot decide}. Each participant evaluated 12 video pairs, including 2 hidden validation pairs with known ground truth.

The validation pairs were constructed by compressing reference videos using high CRF values, producing clearly distinguishable quality differences. These pairs were randomly interleaved with the test samples, and participants were not informed about their presence or purpose. Only responses from participants who correctly answered both validation pairs were retained. To ensure statistical consistency, the comparison graph was balanced such that each video pair received exactly 10 valid annotations. In total, the dataset includes responses from more than 12{,}000 unique participants.

Quality scores were estimated from the pairwise comparisons using the Bradley--Terry model~\cite{bradley1952rank}. The probability that video \( i \) is preferred over video \( j \) is defined as:
\[
P(i \succ j) = \frac{e^{s_i}}{e^{s_i} + e^{s_j}},
\]
where \( s_i \) and \( s_j \) denote the latent subjective quality scores associated with videos \( i \) and \( j \), respectively. These scores are obtained via maximum likelihood estimation based on the observed comparison outcomes.

To assess the reliability of score differences, we compute 95\% confidence intervals for \( s_i - s_j \) under the assumption of asymptotic normality of the maximum likelihood estimates. Let \( \hat{s}_i \) and \( \hat{s}_j \) denote the estimated scores. The variance of their difference is derived from the inverse Fisher information matrix \( I_Y^{-1}(\hat{\theta}) \), where \( \hat{\theta} = \{\hat{s}_1, \hat{s}_2, \ldots\} \).

The standard error is given by:
\[
\hat{\sigma}_{ij} = \sqrt{ \left( I_Y^{-1}(\hat{\theta}) \right)_{ii} + \left( I_Y^{-1}(\hat{\theta}) \right)_{jj} - 2 \left( I_Y^{-1}(\hat{\theta}) \right)_{ij} },
\]
and the corresponding 95\% confidence interval is computed as:
\[
\hat{s}_i - \hat{s}_j \pm 1.96 \, \hat{\sigma}_{ij}.
\]

The use of pairwise comparison is particularly well-suited for large-scale crowdsourcing, as it simplifies the annotation task and improves reliability. Instead of assigning absolute quality scores, participants only need to choose the better of two options, which is cognitively easier and does not require prior training or calibration. This reduces subjectivity and inter-user bias compared to MOS-based protocols, where score interpretation can vary significantly across participants. As a result, pairwise annotations provide more consistent and robust data for estimating perceptual quality rankings.

\section{Benchmarking}
All metrics were evaluated using their publicly available implementations with default settings, without any additional training or tuning, to avoid overfitting and ensure fair comparison. IQA-based methods were applied in a frame-wise manner, where each distorted frame was compared to its reference counterpart and the resulting scores were averaged over the sequence. In contrast, VQA methods directly produce a single quality estimate for the entire video.

Due to the pairwise nature of the subjective annotations, quality scores are only comparable within groups of videos derived from the same reference sequence. Each group includes all distorted versions generated under different codecs and bitrate settings. For every such group, we computed Spearman and Kendall rank correlation coefficients (SROCC and KROCC) between the objective metrics and the corresponding subjective scores. 

To obtain an overall performance measure, the per-group correlations were aggregated using the Fisher Z-transform~\cite{corey1998averaging}, with contributions weighted by group size. The final correlation values were then obtained by applying the inverse Fisher transformation.

\section{Experiments}
In this section, we evaluate objective video quality metrics on the proposed neural video compression dataset. We compare the performance of image- and video- metrics and analyze their alignment with subjective preferences. We also examine how metric behavior differs between NVC and traditional codecs. The following subsections present results on metric correlations and dataset diversity.
\begin{table*}[t]
\caption{Performance of IQA/VQA metrics on the NVC, Traditional, and the whole (General) sets. Best results are \textbf{bold}, second-best are \underline{underlined}. NR stands for the no-reference quality metrics and FR for the full-reference ones.}
\label{tab:kk}
\centering
\setlength{\tabcolsep}{4.2pt}
\small
\begin{tabular}{ll|ccc|ccc|ccc}
\toprule
 & & \multicolumn{3}{c|}{\textbf{NVC}} & \multicolumn{3}{c|}{\textbf{Traditional}} & \multicolumn{3}{c}{\textbf{General}} \\
\cmidrule(lr){3-5} \cmidrule(lr){6-8} \cmidrule(lr){9-11}
\textbf{Type} & \textbf{Metric}
& \textbf{PLCC}$\uparrow$ & \textbf{SROCC}$\uparrow$ & \textbf{KROCC}$\uparrow$
& \textbf{PLCC}$\uparrow$ & \textbf{SROCC}$\uparrow$ & \textbf{KROCC}$\uparrow$
& \textbf{PLCC}$\uparrow$ & \textbf{SROCC}$\uparrow$ & \textbf{KROCC}$\uparrow$ \\
\midrule

\multirow{18}{*}{\textbf{NR}}
& KONIQ++~\cite{su2021koniq++}       & 0.230 & 0.209 & 0.146 & 0.006 & 0.008 & 0.003 & 0.153 & 0.137 & 0.096 \\
& EONSS~\cite{wang2019blind}         & 0.248 & 0.240 & 0.177 & 0.124 & 0.107 & 0.075 & 0.201 & 0.189 & 0.138 \\
& RankIQA~\cite{liu2017rankiqa}      & 0.368 & 0.347 & 0.260 & 0.575 & 0.619 & 0.467 & 0.472 & 0.483 & 0.364 \\
& DOVER~\cite{wu2023exploring}       & 0.371 & 0.349 & 0.245 & 0.898 & \underline{0.874} & \underline{0.724} & 0.642 & 0.616 & 0.498 \\
& LINEARITY~\cite{li2020norm}        & 0.395 & 0.310 & 0.225 & 0.780 & 0.790 & 0.636 & 0.577 & 0.566 & 0.428 \\
& DBCNN~\cite{zhang2020blind}        & 0.439 & 0.284 & 0.186 & 0.719 & 0.753 & 0.599 & 0.579 & 0.523 & 0.384 \\
& KonCept~\cite{hosu2020koniq}       & 0.441 & 0.323 & 0.225 & 0.812 & 0.787 & 0.644 & 0.600 & 0.564 & 0.435 \\
& TOPIQ~\cite{chen2024topiq}         & 0.466 & 0.355 & 0.259 & 0.785 & 0.789 & 0.638 & 0.598 & 0.581 & 0.445 \\
& CLIP-IQA+~\cite{wang2023exploring} & 0.474 & 0.406 & 0.307 & 0.844 & 0.849 & 0.689 & 0.647 & 0.634 & 0.483 \\
& LIQE~\cite{zhang2023blind}         & 0.514 & 0.330 & 0.242 & 0.763 & 0.775 & 0.592 & 0.626 & 0.568 & 0.405 \\
& PIQE~\cite{pandey2020evaluation}   & 0.516 & 0.286 & 0.210 & 0.486 & 0.470 & 0.313 & 0.500 & 0.356 & 0.247 \\
& HyperIQA~\cite{su2020blindly}      & 0.527 & 0.392 & 0.297 & 0.513 & 0.339 & 0.230 & 0.520 & 0.368 & 0.263 \\
& TRES~\cite{golestaneh2022no}       & 0.575 & 0.404 & 0.296 & 0.564 & 0.365 & 0.249 & 0.570 & 0.385 & 0.273 \\
& MDTVSFA~\cite{li2021unified}       & 0.588 & 0.461 & 0.345 & \textbf{0.913} & \textbf{0.883} & \textbf{0.739} & 0.724 & 0.668 & 0.515 \\
& ARNIQA~\cite{agnolucci2024arniqa}  & 0.598 & 0.464 & 0.352 & 0.739 & 0.691 & 0.511 & 0.673 & 0.573 & 0.422 \\
& CNNIQA~\cite{kang2014convolutional}& 0.602 & 0.501 & 0.376 & 0.518 & 0.457 & 0.314 & 0.565 & 0.479 & 0.344 \\
& META-IQA~\cite{Zhu2020MetaIQA}     & 0.627 & 0.442 & 0.331 & 0.861 & 0.835 & 0.676 & 0.733 & 0.623 & 0.470 \\
& MANIQA~\cite{yang2022maniqa}       & 0.685 & 0.481 & 0.354 & 0.781 & 0.745 & 0.564 & 0.731 & 0.612 & 0.446 \\
\midrule

\multirow{8}{*}{\textbf{FR}}
& CONTRIQUE~\cite{madhusudana2022image} & 0.506 & 0.355 & 0.262 & 0.581 & 0.590 & 0.409 & 0.543 & 0.465 & 0.336 \\
& PSNR                                  & 0.576 & 0.516 & 0.399 & 0.806 & 0.793 & 0.647 & 0.680 & 0.649 & 0.523 \\
& TOPIQ~\cite{chen2024topiq}            & 0.559 & 0.538 & 0.418 & 0.842 & 0.821 & 0.666 & 0.695 & 0.676 & 0.538 \\
& SSIM                                  & 0.582 & 0.526 & 0.411 & 0.856 & 0.814 & 0.679 & 0.705 & 0.670 & 0.542 \\
& DISTS~\cite{ding2020image}            & 0.589 & 0.532 & 0.410 & 0.865 & 0.818 & 0.660 & 0.713 & 0.673 & 0.532 \\
& VMAF                                  & 0.593 & 0.528 & 0.411 & 0.878 & 0.858 & 0.708 & 0.722 & 0.702 & 0.559 \\
& LPIPS~\cite{zhang2018unreasonable}      & \underline{0.813} & \textbf{0.629} & \textbf{0.489} & 0.810 & 0.785 & 0.645 & \underline{0.812} & \underline{0.708} & \underline{0.563} \\
& FSIM~\cite{zhang2011fsim}             & \textbf{0.824} & \underline{0.595} & \underline{0.472} & \underline{0.899} & 0.852 & 0.700 & \textbf{0.861} & \textbf{0.724} & \textbf{0.576} \\
\bottomrule
\end{tabular}
\end{table*}
\begin{figure*}[h]
  \centering
  \includegraphics[width=\textwidth]{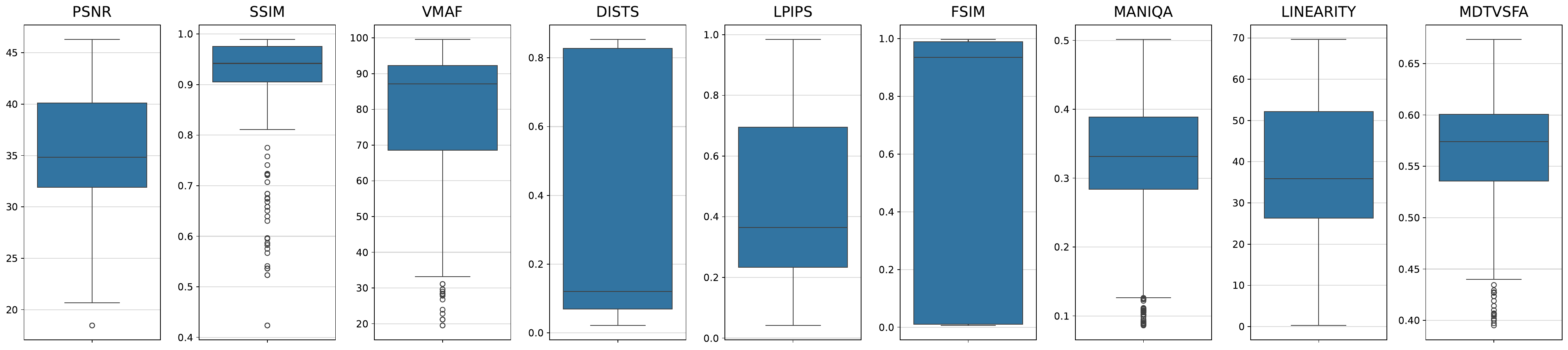}
  \caption{Distribution of metric scores. Each metric appears on a separate axis}
  \label{fig:box}
\end{figure*}

\begin{figure*}[t]
  \centering
  \includegraphics[width=\textwidth]{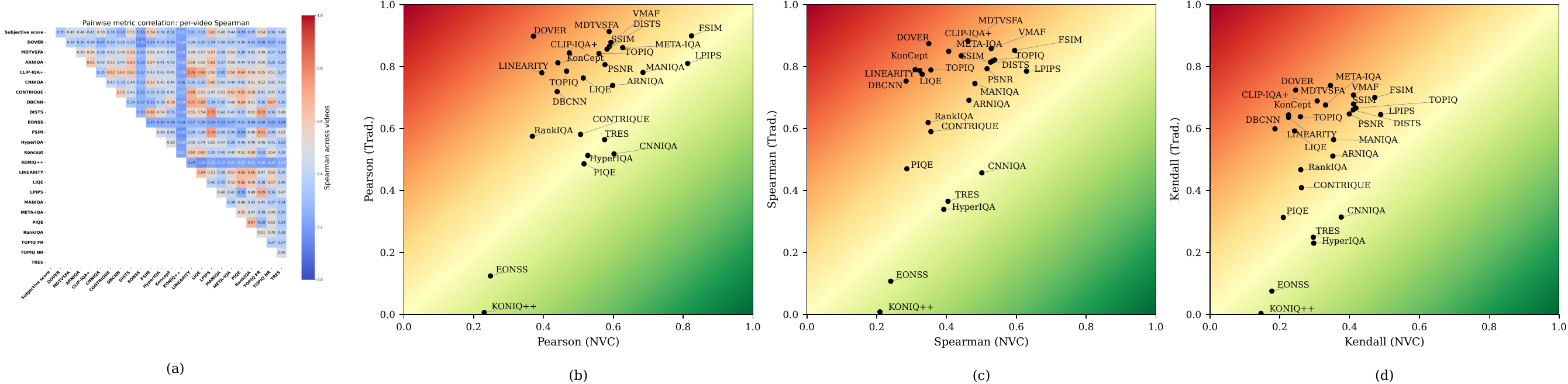}
  \caption{(a) Pairwise Spearman correlation between objective quality metrics computed per video. (b)-(d) Pearson, Spearman and Kendall correlation respectively of objective quality metrics with subjective scores on the proposed dataset NVC and traditional codecs parts. The colormap indicates whether a metric performs better (green area) or worse (red area) on NVC compared to traditional approaches. Most metrics appear above the main diagonal, meaning that they generally perform better for traditional codecs.}
  \label{fig:cors}
\end{figure*}

\subsection{Dataset diversity}
To characterize the diversity and representativeness of the proposed dataset, we performed a comprehensive analysis of spatial and temporal complexity statistics for our dataset, as well as for several widely used public VQA datasets. In particular, we evaluated a set of commonly adopted perceptual and signal-based descriptors, including Spatial Information (SI)~\cite{mackin2018study}, Temporal Information (TI)~\cite{madhusudana2021subjective}, contrast-related measures~\cite{hosu2020koniq}, colorfulness (CF)~\cite{hasler2003measuring}, luminance features~\cite{ying2021patch}, as well as higher-level scoring functions~\cite{chivileva2023measuring} and the IC9600 complexity metric~\cite{feng2022ic9600}. These metrics jointly capture variations in spatial detail, motion intensity, color distribution, and structural complexity, providing a multi-dimensional view of dataset characteristics.

As illustrated in Fig.~\ref{fig:guh}, the proposed dataset demonstrates a broader coverage of the complexity space compared to existing datasets, both in terms of feature range and distribution uniformity. In particular, it exhibits a more balanced distribution across different regions of the feature space, avoiding the concentration effects commonly observed in legacy datasets. Compared to prior compression-oriented benchmarks such as LEHA-CVQAD~\cite{gushchin2025leha}, the proposed dataset shows improved diversity, while also achieving characteristics comparable to large-scale UGC datasets, which are known for their heterogeneous and realistic content. This suggests that the dataset more accurately reflects real-world variability in both content and compression artifacts, making it a suitable benchmark for evaluating modern VQA methods.

\subsection{Objective Metrics Benchmarking}

This study evaluates both recent and state-of-the-art video and image quality metrics for neural video compression (NVC). Since the subjective annotations are obtained via pairwise comparisons of videos derived from the same source sequence, the resulting scores are only comparable within each reference group. Therefore, for each source video, we compute Pearson (PLCC), Spearman (SROCC), and Kendall (KROCC) correlation coefficients between objective metric predictions and subjective scores.

To analyze the behavior of metrics across compression paradigms, we report correlations on three subsets: the full dataset, the NVC subset, and the traditional codecs subset. The results are summarized in Table~\ref{tab:kk}. In most cases, correlations are consistently higher for traditional codecs than for NVC, indicating that existing quality metrics are less aligned with human perception for neural compression artifacts.

Figure~\ref{fig:cors} provides a comparative visualization, where each point corresponds to a metric, with coordinates representing its correlation on NVC (x-axis) and traditional codecs (y-axis). Points lying on the diagonal indicate comparable performance across both domains. Points above the diagonal correspond to better performance on traditional codecs, while points below indicate stronger performance on NVC. Notably, the majority of metrics are concentrated above the diagonal, highlighting a systematic performance gap and suggesting the need for improved or specialized metrics for NVC. Additionally, in the PLCC plot, points appear closer to the diagonal compared to SROCC and KROCC. This suggests that while the relationship between metric predictions and subjective scores remains approximately linear, discrepancies arise in ranking consistency, indicating that current metrics may preserve relative score distances but fail to capture perceptual ordering accurately.

Figure~\ref{fig:cors}(a) presents the pairwise Spearman correlation between objective quality metrics computed on a per-video basis. The results show that while some metrics exhibit moderate agreement, the overall correlation structure is highly heterogeneous. Several clusters of metrics demonstrate relatively strong mutual correlation (e.g., perceptual and learned approaches), whereas others remain weakly correlated, indicating that they capture different aspects of visual quality. Notably, correlations with subjective scores vary significantly across metrics. This behavior highlights the diversity of metric responses and suggests that existing approaches are not fully consistent in their assessment of perceptual quality, particularly in the presence of complex distortions introduced by neural compression.

Figure~\ref{fig:box} shows the distribution of objective quality scores produced by different metrics on the proposed dataset. The results indicate that the metrics operate on substantially different value ranges and exhibit markedly different dispersion patterns. Classical full-reference measures such as PSNR and SSIM produce relatively compact distributions, whereas perceptual metrics such as LPIPS and DISTS span a considerably wider range of values. Learned metrics also demonstrate distinct score distributions, suggesting that they respond differently to the variety of artifacts present in the dataset. Overall, these results confirm that objective metrics are not directly comparable in their raw form and capture different aspects of perceptual quality, which motivates the use of rank-based and correlation-based evaluation in the benchmark.

\section{Conclusion}
In this work, we present a large-scale dataset for neural video compression (NVC) quality assessment. The dataset comprises 2,880 compressed video sequences generated using a diverse set of neural and traditional codecs under various rate–distortion settings. Our experiments demonstrate that widely used video quality assessment metrics exhibit inconsistent performance when applied to neural compression artifacts, highlighting the limitations of existing evaluation approaches. And highlighting the need for dedicated quality metrics specifically designed for neural video codecs. By combining realistic compression scenarios with large-scale subjective annotations, the dataset provides a valuable benchmark for the development and evaluation of VQA models in the context of NVC. We hope this contribution will support future research toward more robust and perceptually aligned video quality metrics for neural compression.

\section{Ethical Considerations and Privacy}

To collect annotations, we employed paid crowdworkers who provided survey responses and expressed quality preferences for pairs of media content. Annotators were compensated at rates exceeding the average local wage for the estimated completion time, reflecting our commitment to fair pay and responsible research practices.

\section*{Acknowledgment}
The work of Nikolay Safonov was supported by the The Ministry of Economic Development of the Russian Federation in accordance with the subsidy agreement (agreement identifier \\ 000000C313925P4H0002; grant No 139-15-2025-012).

The research was carried out using the MSU-270 supercomputer of Lomonosov Moscow State University. The labeling was performed using the Yandex Tasks platform.

\printbibliography

\end{document}